# Multi-constrained Symmetric Nonnegative Latent Factor Analysis for Accurately Representing Large-scale Undirected Weighted Networks

Yurong Zhong, Zhe Xie, Weiling Li, and Xin Luo

**Abstract**—An Undirected Weighted Network (UWN) is frequently encountered in a big-data-related application concerning the complex interactions among numerous nodes, e.g., a protein interaction network from a bioinformatics application. A Symmetric High-Dimensional and Incomplete (SHDI) matrix can smoothly illustrate such an UWN, which contains rich knowledge like node interaction behaviors and local complexes. To extract desired knowledge from an SHDI matrix, an analysis model should carefully consider its symmetric-topology for describing an UWN's intrinsic symmetry. Representation learning to an UWN borrows the success of a pyramid of symmetry-aware models like a Symmetric Nonnegative Matrix Factorization (SNMF) model whose objective function utilizes a sole Latent Factor (LF) matrix for representing SHDI's symmetry rigorously. However, they suffer from the following drawbacks: 1) their computational complexity is high; and 2) their modeling strategy narrows their representation features, making them suffer from low learning ability. Aiming at addressing above critical issues, this paper proposes a Multi-constrained Symmetric Nonnegative Latent-factor-analysis (MSNL) model with two-fold ideas: 1) introducing multi-constraints composed of multiple LF matrices, i.e., inequality and equality ones into a data-density-oriented objective function for precisely representing the intrinsic symmetry of an SHDI matrix with broadened feature space; and 2) implementing an Alternating Direction Method of Multipliers (ADMM)-incorporated learning scheme for precisely solving such a multi-constrained model. Empirical studies on three SHDI matrices from a real bioinformatics or industrial application demonstrate that the proposed MSNL model achieves stronger representation learning ability to an SHDI matrix than state-of-the-art models do.

---◆---

## I. INTRODUCTION

UNDIRECTED Weighted Networks (UWNs) are prevalent in big data applications, particularly protein-related prediction in a protein-protein interaction network [10-14]. Such networks consist of nodes and observed edges, which can be represented as a Symmetric, High-Dimensional, and Incomplete (SHDI) matrix. Note that an SHDI matrix possesses the following characteristics:

1. It is symmetric;
2. Its entities set is large;
3. Its most data entries are missing; and
4. Its data are commonly nonnegative like recommender system's ratings [32].

Despite its incompleteness, an SHDI matrix can provide valuable information, such as identifying potential communities in a social network [7-9]. Therefore, it is essential to design an analysis model that can account for the SHDI matrix's characteristics and extract hidden knowledge.

For effectively represent an SHDI matrix generated from an UWN, recent studies' proposed models can mainly be split into two general categories: 1) Neural Network (NN) models [1, 2]. For example, Sedhain *et al.*'s AutoRec [1] and He *et al.*'s LightGCN [2] are able to extract nonlinear latent factors (LFs) from the target matrix. In spite of attaining nonlinear LFs, they do not consider symmetry and incompleteness of an SHDI matrix; 2) Non-NN models [4, 5, 12, 22, 26, 27]. Nonnegative Matrix Factorization (NMF) models like Xu *et al.*'s NMFC [26] and Leng *et al.*' GNMF [12] are able to well represent the nonnegativity of an SHDI matrix for effectively implementing analysis tasks. However, their modeling do not consider the symmetry and incompleteness of an SHDI matrix. Hence, it is highly desired to propose a pyramid of Symmetric Nonnegative Matrix Factorization (SNMF) models like He *et al.*'s β-SNMF [4], Yang *et al.*'s GSNMF [5] and Hou *et al.*'s PSNMF [27], as well as Nonnegative Latent Factor (NLF) models like Luo *et al*'s NIR [22]. However, SNMF models still does not consider SHDI's incompleteness, while NLF models still cannot represent SHDI's symmetry precisely.

Aiming at addressing above critical issues, this paper proposes a Multi-constrained Symmetric Nonnegative Latent-factor-analysis (MSNL) model. The main contribution of this paper includes:

1. An MSNL model. It introduces multi-constraints into NLF's objective function for precisely representing the intrinsic symmetry

- - - - - - - - - - - - - - - - - - - - - - - - - - - - - - - - - - - - - - - - - -

- *Corresponding author: X. Luo.*
- *Y. Zhong, Z. Xie, W. Li and X. Luo are with the School of Computer Science and Technology, Dongguan University of Technology, Dongguan 523808, China (e-mail: zhongyurong91@gmail.com, gxyz4419@gmail.com, weilinglicq@outlook.com, luoxin21@gmail.com).*

of an SHDI matrix with broadened feature space, and then implements an Alternating Direction Method of Multipliers (ADMM) [26]-incorporated learning scheme for precisely solving such a multi-constrained model;

2. It performs empirical studies on three real-world UWNs to demonstrate that an MSNL model outperforms state-of-the-art models in terms of representation accuracy.

Section II introduces the preliminaries. Section III presents a PSNL model. Section IV gives the experimental results. Finally, Section V concludes this paper.

## II. Preliminaries

### A. Problem Formulation

An SHDI matrix $Y$ generated by an UWN is as defined next.

***Definition 1.*** Given $U$, each entry quantifies some kind of interactions among the matrix $A^{|N| \times |N|}$, which is non-negative. Given known set $\Lambda$ and unknown set $\Gamma$ for $A$, $A$ is an SHDI matrix if $|\Lambda| \ll |\Gamma|$.

For extracting potential yet useful information from an SHDI matrix, an NLF model is defined as:

***Definition 2.*** Given $Y$ and $\Lambda$, an NLF model [6] usually relies on $\Lambda$ to seek for rank-$D$ approximation of $Y$, i.e., $\hat{a}_{m,n} = p_{m,d} * x_{n,d}$ where $p_{m,d} \geq 0$ and $x_{n,d} \geq 0$. With commonly-used Euclidean distance [23, 25, 28] and $L_2$-norm-based regularization scheme [24, 31, 33], the following objective function is defined as:

$$L = \frac{1}{2} \sum_{a_{m,n} \in \Lambda} \left( \left( a_{m,n} - \sum_{d=1}^{D} p_{m,d} x_{n,d} \right)^2 + \frac{\lambda}{2} \sum_{d=1}^{D} \left( \left( p_{m,d} \right)^2 + \left( x_{n,d} \right)^2 \right) \right),$$

$$s.t. \quad \forall m, n \in \{1, 2, \ldots, |N|\}, d \in \{1, 2, \ldots, D\} : p_{m,d} \geq 0, x_{n,d} \geq 0. \tag{1}$$

where the $L_2$-norm-based regularization coefficient $\lambda$ is positive.

## III. An MSNL model

Following the previous study [26], nonnegative constraints applied to the output LFs affect a resultant model's representation accuracy. Hence, we introduce $Q^{|N| \times D}$ and $Y^{|N| \times D}$ into (1) to separate nonnegative constraints from the generalized loss:

$$L = \frac{1}{2} \sum_{a_{m,n} \in \Lambda} \left( \left( a_{m,n} - \sum_{d=1}^{D} q_{m,d} y_{n,d} \right)^2 + \frac{\lambda}{2} \sum_{d=1}^{D} \left( \left( q_{m,d} \right)^2 + \left( y_{n,d} \right)^2 \right) \right),$$

$$s.t. \quad \forall m, n \in \{1, 2, \ldots, |N|\}, d \in \{1, 2, \ldots, D\} : p_{m,d} \geq 0, x_{n,d} \geq 0, \tag{2}$$

$$Q = P, Y = X.$$

Then, for making (2) well describe the symmetry of an SHDI matrix, an equation constraint $P=X$ is introduced into (2), thereby achieving the following symmetry-aware objective function:

$$L = \frac{1}{2} \sum_{a_{m,n} \in \Lambda} \left( \left( a_{m,n} - \sum_{d=1}^{D} q_{m,d} y_{n,d} \right)^2 + \frac{\lambda}{2} \sum_{d=1}^{D} \left( \left( q_{m,d} \right)^2 + \left( y_{n,d} \right)^2 \right) \right),$$

$$s.t. \quad \forall m, n \in \{1, 2, \ldots, |N|\}, d \in \{1, 2, \ldots, D\} : p_{m,d} \geq 0, x_{n,d} \geq 0, \tag{3}$$

$$Q = P, Y = X, P = X,$$

For solving MSNL's objective function (3) efficiently, its learning scheme is designed by following the principle of ADMM. Firstly, (3) should be reformulated as the following augmented Lagrangian function:

$$\varepsilon = \frac{1}{2} \sum_{a_{m,n} \in \Lambda} \left( \left( a_{m,n} - \sum_{d=1}^{D} q_{m,d} y_{n,d} \right)^2 + \frac{\lambda}{2} \sum_{d=1}^{D} \left( \left( q_{m,d} \right)^2 + \left( y_{n,d} \right)^2 \right) \right)$$

$$+ \sum_{j=1}^{|N|} \sum_{d=1}^{D} u_{j,d} \left( q_{j,d} - p_{j,d} \right) + \sum_{j=1}^{|N|} \sum_{d=1}^{D} \frac{\theta_1}{2} \left( q_{j,d} - p_{j,d} \right)^2$$

$$+ \sum_{j=1}^{|N|} \sum_{d=1}^{D} v_{j,d} \left( y_{j,d} - x_{j,d} \right) + \sum_{j=1}^{|N|} \sum_{d=1}^{D} \frac{\theta_1}{2} \left( y_{j,d} - x_{j,d} \right)^2$$

$$+ \sum_{j=1}^{|N|} \sum_{d=1}^{D} w_{j,d} \left( p_{j,d} - x_{j,d} \right) + \sum_{j=1}^{|N|} \sum_{d=1}^{D} \frac{\theta_2}{2} \left( p_{j,d} - x_{j,d} \right)^2, \tag{4}$$

$$s.t. \quad \forall j \in \{1, 2, \ldots, |N|\}, d \in \{1, 2, \ldots, D\} : p_{j,d} \geq 0, x_{j,d} \geq 0$$

Note that $\theta_1$ and $\theta_2$ are set as $\theta_1 = \beta_1 |\Lambda(j)|$ and $\theta_2 = \beta_2 |\Lambda(j)|$, respectively, where $\beta_1$ and $\beta_2$ are positive constants.

Then, according to solution algorithms for the augmented Lagrangian function (4) [3, 26], $\forall m, n \in \{1, 2, \ldots, |N|\}$, $d \in \{1, 2, \ldots, D\}$, the learning rules of optimized parameters in $Q$, $Y$, $P$, $X$, $U$, $V$ and $W$ are given as:

$$q_{m,d} = \frac{\sum\limits_{n \in \Lambda(m)} \left( a_{m,n} - \sum\limits_{l=1, l \neq d}^{D} q_{m,l} y_{n,l} \right) y_{n,d} + \theta_1 p_{m,d} - u_{m,d}}{\sum\limits_{n \in \Lambda(m)} \left( \left( y_{n,d} \right)^2 + \lambda \right) + \theta_1}, \tag{5a}$$

$$y_{m,d} = \frac{\sum\limits_{n \in \Lambda(m)} \left( a_{n,m} - \sum\limits_{l=1, l \neq d}^{D} q_{n,l} y_{m,l} \right) q_{n,d} + \theta_1 x_{m,d} - v_{m,d}}{\sum\limits_{n \in \Lambda(m)} \left( \left( q_{n,d} \right)^2 + \lambda \right) + \theta_1}, \tag{5b}$$

$$p_{m,d} = max \left( 0, \frac{\theta_1 q_{m,d} + \theta_2 x_{m,d} + u_{m,d} - w_{m,d}}{\theta_1 + \theta_2} \right), \tag{5c}$$

$$x_{m,d} = max \left( 0, \frac{\theta_1 y_{m,d} + \theta_2 p_{m,d} + v_{m,d} + w_{m,d}}{\theta_1 + \theta_2} \right), \tag{5d}$$

$$u_{m,d} \leftarrow u_{m,d} + \eta \theta_1 \left( q_{m,d} - p_{m,d} \right), \tag{5e}$$

$$v_{m,d} \leftarrow v_{m,d} + \eta \theta_1 \left( y_{m,d} - x_{m,d} \right), \tag{5f}$$

$$w_{m,d} \leftarrow w_{m,d} + \eta \theta_2 \left( p_{m,d} - x_{m,d} \right), \tag{5g}$$

where (5a)-(5d) are achieved by element-wise alternating least square algorithm, (5e)-(5g) are achieved by the dual gradient ascent algorithm, and nonnegative truncation method is used to guarantee the nonnegativity of $p_{m,d}$ and $x_{m,d}$.

Afterward, MSNL's whole optimization task described by (5) is split into $D$ disjoint subtasks where each subtask contains three jobs corresponding to a specific <u>L</u>atent <u>F</u>actor (LF) dimension $d$, i.e., $\forall d \in \{1 \sim D\}$, the $d$-th task consists of the following jobs:

**1. Job One:**

$$\begin{cases} Q_{:,d}^{i+1} \overset{(5a)}{\leftarrow} \underset{Q_{:,d}}{\arg \min} \, \varepsilon \left( \begin{bmatrix} Q_{:,1 \sim (d-1)}^{i+1}, Q_{:,d}^i, Q_{:,(d+1) \sim D}^i \end{bmatrix}, \begin{bmatrix} Y_{:,1 \sim (d-1)}^{i+1}, Y_{:,d \sim D}^i \end{bmatrix}, \\ P_{:,d}^i, X_{:,d}^i, U_{:,d}^i, V_{:,d}^i, W_{:,d}^i \end{bmatrix} \right), \\ Y_{:,d}^{i+1} \overset{(5b)}{\leftarrow} \underset{X_{:,d}}{\arg \min} \, \varepsilon \left( \begin{bmatrix} Q_{:,1 \sim (d-1)}^{i+1}, Q_{:,d \sim D}^i \end{bmatrix}, \begin{bmatrix} Y_{:,1 \sim (d-1)}^{i+1}, Y_{:,d}, Y_{:,(d+1) \sim D}^i \end{bmatrix}, \\ P_{:,d}^i, X_{:,d}^i, U_{:,d}^i, V_{:,d}^i, W_{:,d}^i \end{bmatrix} \right), \end{cases} \tag{6a}$$

**2. Job Two:**

$$\begin{cases} P_{:,d}^{i+1} \overset{(5c)}{\leftarrow} \underset{P_{:,d} \geq 0}{\arg \min} \, \varepsilon \left( Q_{:,d}^{i+1}, Y_{:,d}^{i+1}, P_{:,d}, X_{:,d}^i, U_{:,d}^i, V_{:,d}^i, W_{:,d}^i \right), \\ X_{:,d}^{i+1} \overset{(5d)}{\leftarrow} \underset{Y_{:,d} \geq 0}{\arg \min} \, \varepsilon \left( Q_{:,d}^{i+1}, Y_{:,d}^{i+1}, P_{:,d}^i, X_{:,d}, U_{:,d}^i, V_{:,d}^i, W_{:,d}^i \right), \end{cases} \tag{6b}$$

**3. Job Three:**

$$\begin{cases} U_{:,d}^i \overset{(5e)}{\leftarrow} U_{:,d}^i + \eta \theta_1 \nabla_U \, \varepsilon \left( Q_{:,d}^{i+1}, Y_{:,d}^{i+1}, P_{:,d}^{i+1}, X_{:,d}^{i+1}, U_{:,d}^i, V_{:,d}^i, W_{:,d}^i \right), \\ V_{:,d}^{i+1} \overset{(5f)}{\leftarrow} V_{:,d}^i + \eta \theta_1 \nabla_V \, \varepsilon \left( Q_{:,d}^{i+1}, Y_{:,d}^i, P_{:,d}^{i+1}, X_{:,d}^{i+1}, U_{:,d}^i, V_{:,d}^i, W_{:,d}^i \right), \\ W_{:,d}^{i+1} \overset{(5g)}{\leftarrow} W_{:,d}^i + \eta \theta_2 \nabla_W \, \varepsilon \left( Q_{:,d}^{i+1}, Y_{:,d}^i, P_{:,d}^{i+1}, X_{:,d}^{i+1}, U_{:,d}^i, V_{:,d}^i, W_{:,d}^i \right), \end{cases} \tag{6c}$$

where (6) is designed with the following considerations: 1) each subtask deal with the update of optimization parameter related to specific LF dimension, i.e., the $d$-th column of $Q$, $Y$, $P$, $X$, $U$, $V$ and $W$. Hence, information hidden in $\Lambda$ can be used totally; 2) each subtask related to $Q$ or $Y$ is solved based on the solution to those solved before. Moreover, Job Two and Three address parameters in the $d$-th column of $P$, $X$, $U$, $V$ and $W$ following a standard ADMM process [3, 26].

## IV. EXPERIMENTAL RESULTS AND ANALYSIS

### A. General Settings

**Evaluation Protocol.** In real-world applications, decomposing an SHDI matrix into LFs is critical for predicting missing values and identifying potential connections between entities [10, 11]. As a result, this technique is often employed as an evaluation protocol for assessing the performance of related models.

**Evaluation Metrics.** The accuracy of a test model for missing data predictions can be measured by the root mean square error (RMSE) [15-18, 30]:

$$RMSE = \sqrt{\left(\sum_{r_{i,j} \in \Gamma} (r_{i,j} - \hat{r}_{i,j})^2\right) \Big/ |\Gamma|},$$

where $\Gamma$ denotes the validation set and is disjoint with the training set $\Lambda$. Note that low RMSE represents high prediction accuracy for missing data in $\Gamma$.

**Datasets.** Our experiments adopt three UWNs, and their details are shown in Table I. In all experiments on each dataset, we randomly split its known set of entries $\Lambda$ into ten disjoint subsets for tenfold cross-validation. We adopt seven subsets as the training set, one subset as the validation set, and the remaining two subsets as the test set. This process is repeated ten times sequentially.

TABLE I. Details of Adopted Datasets.

| No. | Type | $|\Lambda|$ | $|N|$ | Density | Source |
|-----|------|-----|-----|---------|--------|
| D1 | Protein | 1,182,124 | 5,194 | 4.38% | [19] |
| D2 | Protein | 1,120,028 | 7,963 | 1.77% | [19] |
| D3 | Material | 322,905 | 13,965 | 0.17% | [20] |

TABLE II. Details of Tested Models.

| No. | Name | Description |
|-----|------|-------------|
| M1 | NMFC | A commonly-adopted ADMM-incorporated NMF model [26]. |
| M2 | NIR | A recent NLF model [22]. |
| M3 | $\beta$-SNMF | A commonly-adopted SNMF model [4]. |
| M4 | GSNMF | A commonly-adopted SNMF model [5]. |
| M5 | LightGCN | A commonly-adopted Graph Convolutional Network model [2]. |
| M6 | PSNL | The proposed model in this paper. |

**Compared Models.** Our experiments involve six models, and their details are presented in Table II. To achieve its objective results, we used the following settings:

(1) LF Dimension $d$ is set at 20;

(2) For each model on each data set, the results generated from 10 different random initial values are recorded to calculate the average RMSE and convergence time for eliminating the effect of initial assumptions [21, 29].

(3) The training process of the test model is terminated when: 1) the iteration count reaches a preset threshold, which is 1000; 2) The difference between the two consecutive iterations of the generated RMSE is less than $10^{-5}$.

*B. Comparison against State-of-the-art Models*

TABLE III. RMSE of M1-6 on D1-3.

| No. | M1 | M2 | M3 | M4 | M5 | M6 |
|-----|-----|-----|-----|-----|-----|-----|
| D1 | 0.1437$_{\pm 2.0E-4}$ | 0.1401$_{\pm 1.5E-4}$ | 0.1806$_{\pm 4.1E-4}$ | 0.1972$_{\pm 1.4E-4}$ | 0.1415$_{\pm 1.3E-4}$ | **0.1384$_{\pm 2.3E-4}$** |
| D2 | 0.1329$_{\pm 2.8E-4}$ | 0.1276$_{\pm 1.6E-4}$ | 0.1650$_{\pm 2.7E-4}$ | 0.1813$_{\pm 1.6E-4}$ | 0.1297$_{\pm 1.4E-4}$ | **0.1264$_{\pm 3.3E-4}$** |
| D3 | 0.0820$_{\pm 1.9E-5}$ | 0.0752$_{\pm 2.0E-4}$ | 0.0739$_{\pm 1.2E-5}$ | 0.0764$_{\pm 8.1E-5}$ | 0.1037$_{\pm 2.2E-4}$ | **0.0720$_{\pm 4.9E-4}$** |

TABLE IV. Total Time Cost of M1-6 on D1-3 (Seconds).

| No. | M1 | M2 | M3 | M4 | M5 | M6 |
|-----|-----|-----|-----|-----|-----|-----|
| D1 | 1,103$_{\pm 74.94}$ | 68$_{\pm 8.29}$ | 2,101$_{\pm 247.69}$ | 2,477$_{\pm 231.54}$ | 1,298$_{\pm 135.77}$ | 379$_{\pm 24.19}$ |
| D2 | 2,389$_{\pm 185.34}$ | 53$_{\pm 6.74}$ | 5,107$_{\pm 428.91}$ | 9,542$_{\pm 980.55}$ | 1,106$_{\pm 122.48}$ | 269$_{\pm 27.23}$ |
| D3 | 5,251$_{\pm 451.69}$ | 118$_{\pm 17.86}$ | 852$_{\pm 76.98}$ | 2,631$_{\pm 326.11}$ | 94$_{\pm 10.91}$ | 80$_{\pm 14.76}$ |

Table III and IV summarize RMSE and total time cost of M1-6 on D1-3, respectively. From these results, we have the following findings:

(1) **MSNL achieves higher accuracy gain than state-of-the-art models do.** For example, as shown in Table III, RMSE of M6 is 0.0720 on D3, which is about 12.2%, 4.26%, 2.57%, 5.76% and 30.57% lower than M1's 0.0820, M2's 0.0752, M3's 0.0739, M4's 0.0764 and M5's 0.1037, respectively. Similar results can be found on D1 and D2.

(2) **MSNL's computational efficiency is competitive.** MSNL's total time cost is the least on D3, as recorded in Table IV. Hence, MSNL's computational efficiency is competitive.

## V. CONCLUSIONS

An MSNL model has shown great potential in predicting missing data of an SHDI matrix generated by an UWN. Its highly accurate representation boosts its practicability. In the future, we will continue to explore parallelization mechanism for the model's learning algorithm, thereby further improving the model's computational efficiency.